\documentclass[letterpaper, 10 pt, conference]{ieeeconf}  

\IEEEoverridecommandlockouts                              

\title{\LARGE \bf
PA-BiCoop: A Primary-Auxiliary Cooperative Framework for General Bimanual Manipulation
}

\author{Qicheng Bai$^{1}$, Ziru Wang$^{1}$, Teli Ma$^{2}$, Guang Dai$^{1}$, Jingdong Wang$^{3}$ and Mengmeng Wang$^{4,1,*}$ 
\thanks{This work was supported by Zhejiang Province Natural Science Foundation of China under Grant
No. LQN25F030008 and the National Natural Science Foundation of China under Grant No. 62403429.}
\thanks{$^{1}$SGIT AI Lab, State Grid Corporation of China.}
\thanks{$^{2}$The Hong Kong University of Science
and Technology, Guangzhou.}
\thanks{$^{3}$Baidu Research, Beijing, China.}
\thanks{$^{4}$Zhejiang
University of Technology, Hangzhou.}
\thanks{$^{*}$Corresponding Author.}
}

\usepackage[T1]{fontenc}
\usepackage{graphicx}
\usepackage{cite}
\begin{document}
\bibliographystyle{bst/IEEEtran}

\maketitle
\thispagestyle{empty}
\pagestyle{empty}
\newcommand{\name}{PA-BiCoop~}
\newcommand{\nameo}{PA-BiCoop}


\begin{abstract}
Bimanual manipulation is essential for advanced robotic systems because it offers higher efficiency and flexibility compared to single-arm configurations. However, existing approaches either lack inter-arm interaction or ignore the need for a dynamic division of labor, treating the arms as functionally equivalent.
To address these limitations, this paper draws inspiration from human bimanual manipulation where one arm handles core operations and the other provides auxiliary support, and proposes \nameo, a new single-model bimanual cooperation framework with dynamic primary-auxiliary arm differentiation. \name categorizes robotic arms into primary and auxiliary arms with adaptively adjustable roles across task stages, employs two specialized decoders that share a global feature encoder: the primary decoder generates the primary arm’s base-coordinate pose and core-task affordance heatmaps, and the auxiliary decoder outputs the auxiliary arm’s relative pose in the primary arm’s coordinate system. Moreover, we design a dynamic role assignment module to automatically map roles to left/right arms without manual pre-definition. This design facilitates inter-arm knowledge sharing and coordinated manipulation.
Extensive experiments demonstrate that our \name achieves superior performance: it outperforms state-of-the-art baselines by 48\% on average in RLBench2 simulation tasks and by over 50\% on average in real-world tasks, thereby verifying its effectiveness and advancement in bimanual manipulation.

\end{abstract}


\section{Introduction}

Bimanual manipulation \cite{nakamura1989dynamics, paljug2002control, sarkar1997dynamic, smith2012dual} has become an indispensable component in advanced robotic systems, owing to its superior efficiency and operational flexibility compared to unilateral (single-arm) configurations~\cite{goyal2023rvt, shridhar2023perceiver, fangsam2act, goyal2024rvt, brohan2022rt}. Notably, it enables the execution of tasks that are inherently unattainable for single-arm setups, such as transporting large objects, removing bottle caps, or assembling complex components. Compared to single-arm manipulation, bimanual operation poses significantly greater challenges, as it requires simultaneously modeling the states and actions of both arms, along with their collaborative interactions.

\begin{figure}[t]
\centering
\includegraphics[width=0.5\textwidth]{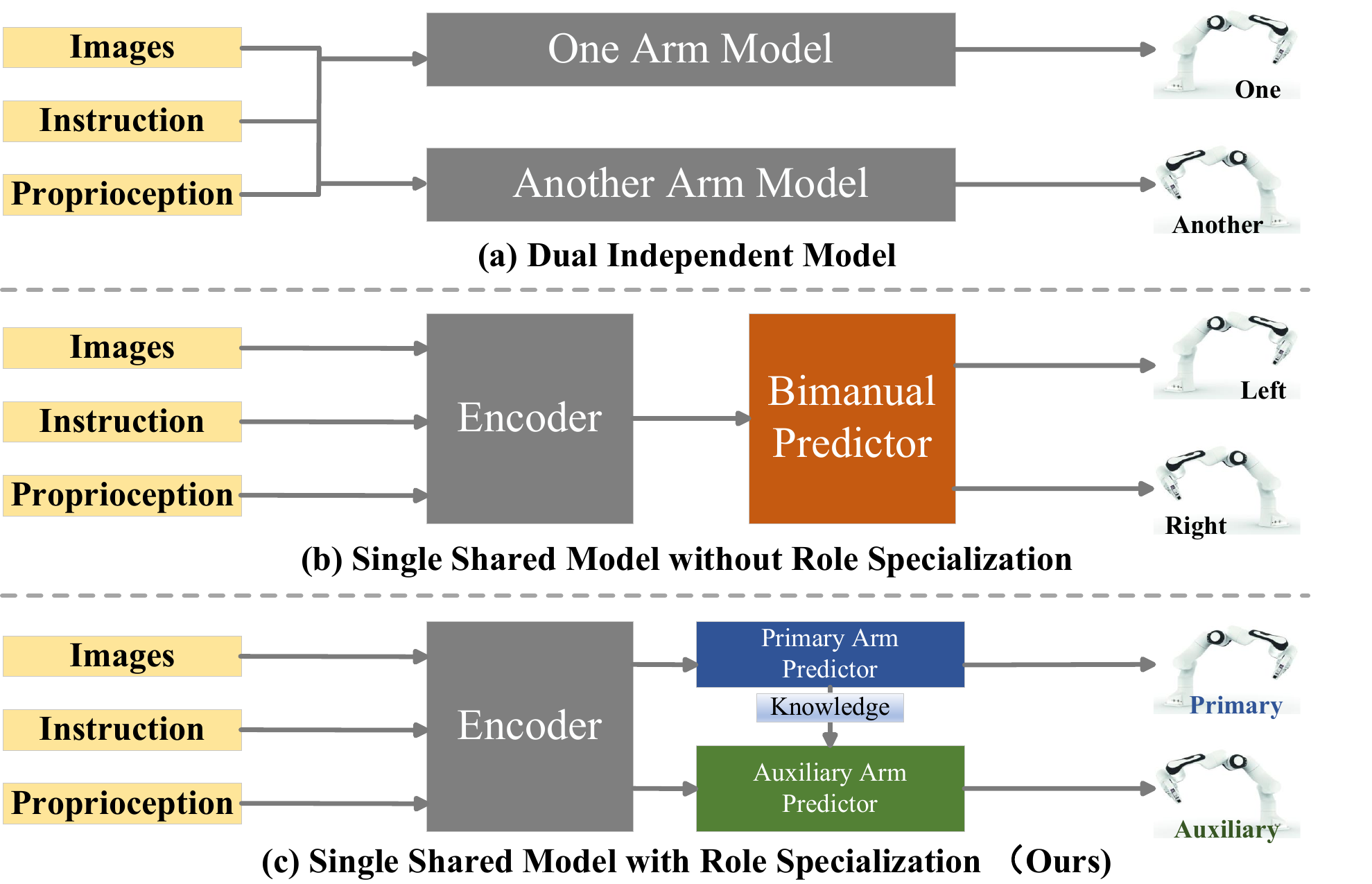} 
\caption{Three model paradigms for bimanual manipulation.}
\label{fig1}
\vspace{-0.50cm}
\end{figure}

As shown in Fig. \ref{fig1}, the predominant approaches in this domain can be categorized into two architectural paradigms currently. The first employs dual independent models to predict the movements of each arm separately, with representative methods including AnyBimanual \cite{Lu2025anybimanual}, VoxActB \cite{liu2025voxact}, and BUDS \cite{grannen2023stabilize}. While these approaches draw extensively on the research findings and technical advancements in single-arm modeling, they essentially adopt an independent modeling framework where each arm is assigned its own dedicated model. Therefore, this paradigm has a core limitation: it lacks inter-arm knowledge sharing and interactive information transfer, thereby compromising overall task performance. Additionally, the use of two separate models inevitably doubles the model complexity during both training and inference phases.
In contrast, the second paradigm adopts a single shared modeling framework for the joint modeling of both arms, as exemplified by YOTO \cite{zhou2025you}, PerAct2 \cite{grotz2024peract2}, and Kstar Diffuser \cite{Lv_2025_CVPR}. However, it can be observed that despite leveraging a single shared model, these methods usually treat the left and right arms as functionally equivalent without distinguishing their roles. They generate movements for the two arms in one of two rigid modes: either asynchronously following a manually predefined sequence \cite{zhou2025you} or simultaneously outputting the action spaces of both arms \cite{grotz2024peract2, Lv_2025_CVPR, liu2024rdt, zhaolearning}. 

To illustrate this limitation, this paper draws inspiration from human bimanual manipulation: when humans perform tasks such as assembling parts, wrapping packages, or using tools, their two arms rarely act in a "role-agnostic" manner. Instead, one primary arm typically takes charge of core operations. Examples include aligning components or wielding a tool. Meanwhile, the other auxiliary arm provides auxiliary support, such as stabilizing the workpiece or handing over materials. Crucially, this role division is not fixed, as the left and right arms can dynamically switch roles depending on the task demands. This inherent division of labor and adaptive collaboration is precisely what is missing in the aforementioned shared-model paradigms, ultimately leading to suboptimal coordination efficiency in complex bimanual tasks.

Motivated by this human bimanual manipulation mechanism, we introduce \textbf{PA-BiCoop}, a new single-model \textbf{Bi}manual \textbf{Coop}eration framework for bimanual manipulation that incorporates \textbf{P}rimary-\textbf{A}uxiliary differentiation. Unlike previous single-model frameworks, our approach dynamically assigns distinct roles and collaborative functions to the two robotic arms as shown in Fig. \ref{fig1} (c): we categorize the robotic arms as primary arm and auxiliary arm, with the distinction being non-fixed and adaptively adjusted between the two arms at different task stages. For the primary arm, we employ a primary decoder to generate main affordance heatmaps indicating the current primary task region and output the primary arm’s pose under the base coordinate system. For the auxiliary arm, we propose an auxiliary decoder that outputs the auxiliary arm’s relative pose in the coordinate system of the primary arm based on the primary arm’s affordance heatmaps and global features. These two decoders share a global feature encoder, allowing for the sharing of global information without the need for redundant perception models and thus maintaining the simplicity of the overall architecture.
This design of primary and auxiliary decoders facilitates inter-arm knowledge sharing. Furthermore, we design a role assignment module that dynamically maps the primary and auxiliary roles to the left and right arms, eliminating the need for pre-defined manual role sequences. Evaluation across both simulation tasks and real-world tasks demonstrates the effectiveness of our \nameo, which achieves substantial advancements over prior methods. 

Our contributions can be summarized as follows:
\begin{itemize}
    \item We propose \nameo, a new single-model bimanual framework that enables dynamic primary-auxiliary specialization, enhancing collaboration through role-aware design.
\item We introduce dedicated primary and auxiliary decoders for action prediction and inter-arm interaction, along with a learnable role assignment module enabling automatic role specialization.
\item Extensive experiments show that \name outperforms state-of-the-art methods by 48\% on RLBench2~\cite{grotz2024peract2} and over 50\% in real-world tasks, achieving superior overall success rates.
\end{itemize}
\section{Related Work}
Current methodologies in bimanual manipulation can be broadly categorized into two architectural paradigms: dual independent models and single shared models.

\textbf{Dual-Model Architecture.}
Several approaches mitigate coordination challenges through dual-model architectures \cite{sirintuna2023carrying, liu2025voxact, Lu2025anybimanual, gao2024bi, liu2022robot, zhou2016coordinate, ijspeert2013dynamical, zhou2016learning, saveriano2023dynamic}. Early methods \cite{liu2022robot, zhou2016coordinate, ijspeert2013dynamical, zhou2016learning, saveriano2023dynamic, zhang2021dair} propose to create a “leader and follower” movement, suffering from large memory consumption and fixed roles. Methods such as BUDS \cite{grannen2023stabilize} and VoxActb \cite{liu2025voxact} decouple the system into stabilizing and acting arms. VoxActb \cite{liu2025voxact} employs VLMs \cite{achiam2023gpt,touvron2023llama} for scene region prioritization and voxel grid reconstruction. However, they lack flexible role switching between arms and are limited to relatively simple tasks, constraining their performance in complex scenarios. AnyBimanual \cite{Lu2025anybimanual} proposes a model-agnostic plug-and-play framework that generalizes pretrained single-arm policies \cite{shridhar2023perceiver, goyal2023rvt}. Although it uses attention partitioning to isolate arm-specific regions of interest, this design also restricts knowledge sharing between arms. Overall, these methods underutilize inter-arm interaction, limiting their effectiveness in tasks requiring high coordination.

\textbf{Single-Model Architecture.}
Alternative approaches employ a single-model framework \cite{liu2024rdt, Lv_2025_CVPR, zhou2025you, grotz2024peract2, zhaolearning, fu2024mobile, lee2025interact, yu2024bikc, chen2023bi, gbagbe2024bi, kataoka2022bi, caccavale2017imitation, shi2023waypoint} to leverage inter-arm knowledge sharing. Early work such as ACT \cite{zhaolearning} uses conditional VAEs \cite{kingma2013auto} within an encoder-decoder structure for joint angle prediction. PerAct2 \cite{grotz2024peract2} extends PerAct \cite{shridhar2023perceiver} by duplicating its prediction head for simultaneous dual-arm control. InterACT \cite{lee2025interact} effectively captures dependencies between joint states and visual inputs through a hierarchical attention mechanism. KStar Diffuser \cite{Lv_2025_CVPR} incorporates spatiotemporal graphs and differentiable kinematics to guide diffusion models. A common limitation among these methods is their treatment of both arms as functionally equivalent without explicit role specialization, thereby overlooking inherent asymmetries in task division. YOTO \cite{zhou2025you} employs a single prediction head for bimanual action output but depends on handcrafted coordination sequences, resulting in poor synchronous performance. Although unified in structure, these methods generally fail to explicitly model bimanual coordination, making them susceptible to failure from minor motion discrepancies.

In this paper, we propose a new approach that incorporates dynamic primary–auxiliary arm differentiation within a single shared model. This design enhances cross-arm knowledge sharing and interaction, facilitating high-precision coordination in complex bimanual tasks.
\begin{figure*}[t]
\centering
\includegraphics[width=1.0\textwidth]{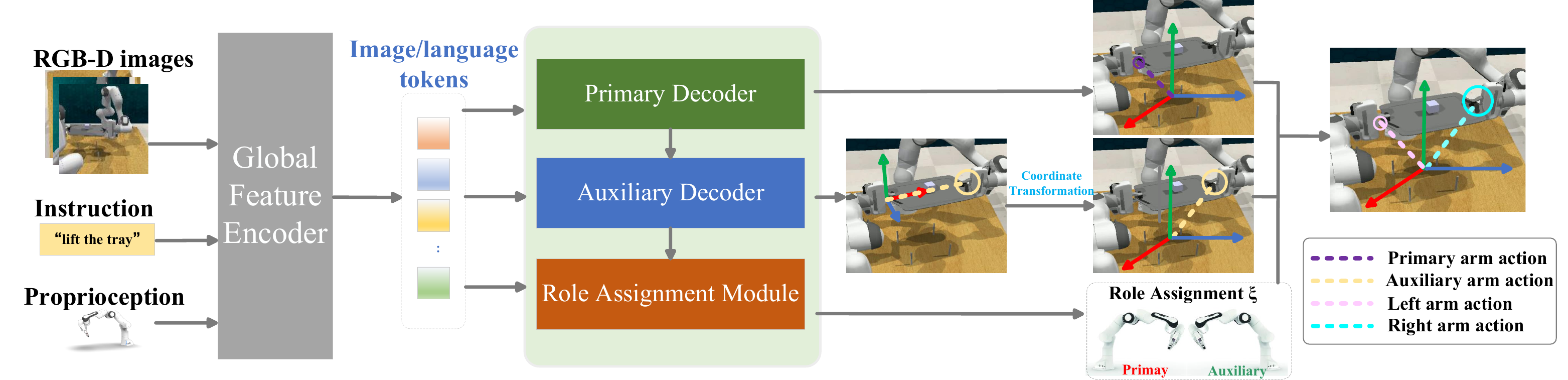} 
\caption{The framework of \nameo. Given RGB-D images, instruction, and proprioception, we encode them through a global feature encoder to image/language tokens. The primary decoder, auxiliary decoder, and role assignment module processes these tokens to produce the primary action $a^P_{bc}$, the auxiliary action $a^A_{pc}$, and the role assignment $\xi$. $a^A_{pc}$ is transformed to $a^A_{bc}$ through coordinate transformation. $\xi$ maps primary/auxiliary arms to left/right arms.}
\label{fig2}
\end{figure*}

\section{Method}
\subsection{Problem Formulation}


Our goal is to learn a model $a=f_\theta(o,l,p)$ that can complete various bimanual manipulation tasks, where $o$ contains RGB-D images from multiple perspectives, $l$ is the language description of the task, $p$ is the proprioception of two robot arms, and the output $a$ is the actions of two robot arms. The action of each arm consists of the 6-DoF end-effector pose (3-DoF for translation and 3-DoF for rotation), 1-DoF gripper state (open or close), and a binary indicator for whether to allow collision avoidance for the motion planner. In addition, referring to prior works \cite{shridhar2023perceiver, goyal2023rvt, grotz2024peract2, ma2024hierarchical}, we employed key-frame extraction technology to enhance learning efficiency. Thus, the model only needs to predict action at the next key-frame.

\subsection{Framework}
Inspired by the inherent division of labor in human bimanual manipulation, where one arm typically takes the lead in core tasks while the other provides auxiliary support, we analogously categorize the robotic arms in our system into \textbf{Primary Arm} and \textbf{Auxiliary Arm}. 
The general pipeline of our proposed \name is shown in Fig. \ref{fig2}. Considering both the efficiency of feature extraction and the effectiveness of global feature representation~\cite{goyal2023rvt, shridhar2023perceiver, fangsam2act, zhu2023viola}, the RVT \cite{goyal2023rvt} is employed as the global shared feature encoder, which processes $o$, $l$, and $p$ to produce image tokens and language tokens. These tokens are subsequently decoded by the primary decoder, auxiliary decoder, and role assignment module to output the primary arm action $a^P_{bc}$, the auxiliary arm action $a^A_{pc}$, and the role assignment variable $\xi$ respectively. 

The primary arm's action $a^P_{bc}$ is generated in the base coordinate system $C_{bc}$ using the global feature representation, leveraging its precise multi-view perception capability. The auxiliary arm's action $a^A_{pc}$ is predicted in the primary arm's coordinate system $C_{pc}$, dependent on the primary arm's kinematic state. This approach exploits the relative positioning between arms during bimanual coordination tasks, reducing the burden on the auxiliary arm's spatial perception and the complexity of coordination. The role assignment module predicts $\xi$ to enable dynamic role switching: $\xi[0]$=$1$ designates the left arm as the primary arm (right as the auxiliary arm), while $\xi[1]$=$0$ reverses this mapping, allowing context-dependent role switching. The resulting $a^A_{pc}$ is transformed to the action $a^A_{bc}$ in the coordinate system $C_{bc}$ using coordinate transformation. 
Subsequently, we will provide a detailed explanation of the primary decoder, the auxiliary decoder, the role assignment module, the coordinate transformation, and the training loss functions.
\subsection{Primary Decoder}
\begin{figure*}[t]
\centering
\includegraphics[width=1.0\textwidth]{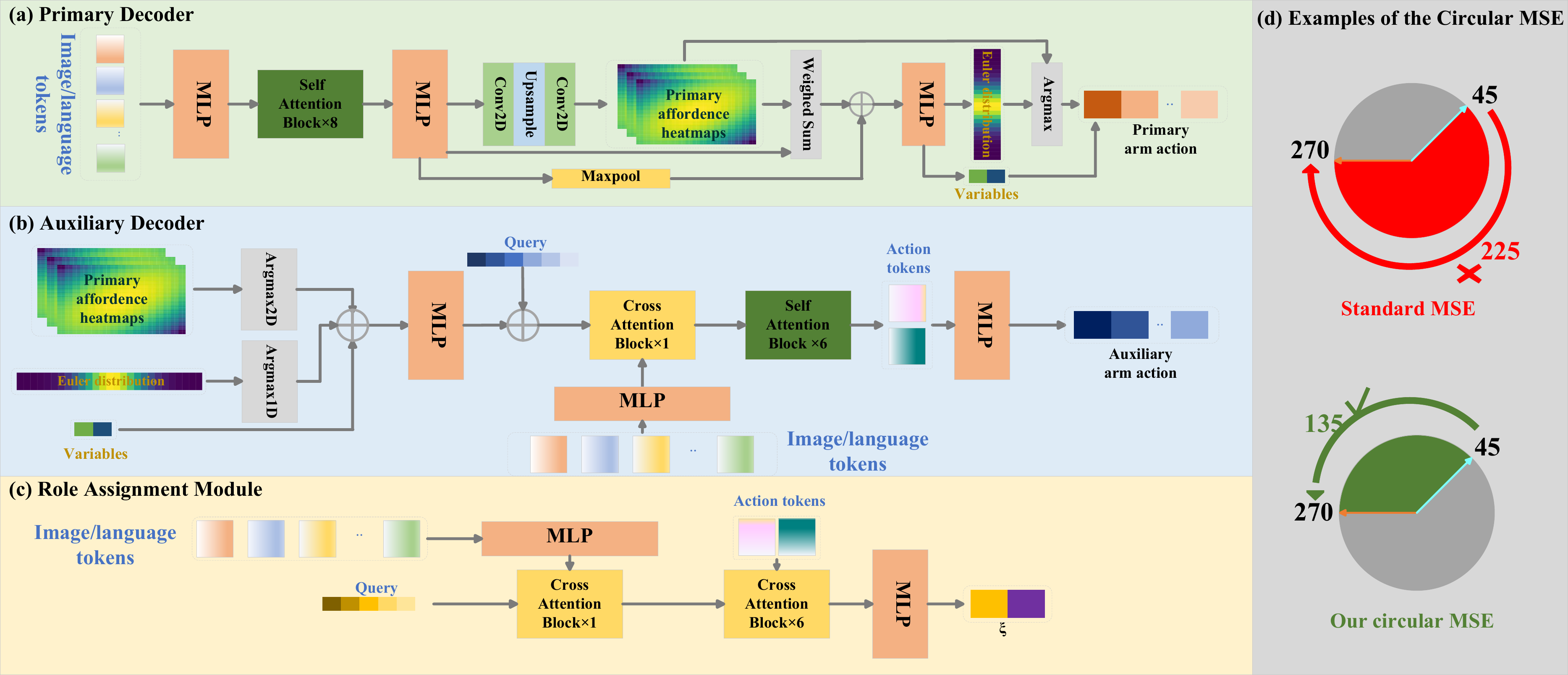} 
\caption{(a) \textbf{The primary decoder.} It primarily employs self-attention blocks, convolutional layers, and MLPs to generate the main affordance heatmaps and ultimately predict actions for the primary arm based on global image/language tokens. (b) \textbf{The auxiliary decoder.} This component consists mainly of cross-attention blocks, self-attention blocks, and MLPs, which utilize outputs from the primary decoder along with global image/language tokens to predict actions for the auxiliary arm. (c) \textbf{The role assignment module.} Comprising cross-attention blocks and MLPs, this module determines the functional role specialization of the two robotic arms based on the global image/language tokens and the action tokens from the auxiliary decoder. (d) \textbf{Examples of the circular MSE.} The circular MSE correctly computes the angular distance between $45^\circ$ and $270^\circ$ (green arc), whereas standard MSE yields an invalid result (red arc).}

\label{fig3}
\vspace{-0.50cm}
\end{figure*}

As shown in Fig.~\ref{fig3}~(a), the primary decoder uses the global features to generate affordance heatmaps and predict the primary arm's actions. First, we feed image and language tokens into a multi-layer perceptron (MLP) to adjust features. Subsequently, we utilize eight transformer layers (each comprising multi-head self-attention and an MLP) to focus on the task region and output image features across different views. The image features are then processed through two convolutional layers to generate the main affordance heatmaps $H^P_T$, which indicate the current primary task region for the primary arm across different views, with peak values corresponding to the desired translation coordinates of the primary arm. To meet the accuracy requirements for primary arm prediction in image-patched tokens, we adopt bilinear interpolation \cite{kirkland2010bilinear} to upsample multi-view tokens to their original spatial resolution between convolutional layers. For rotation and other action variables, we use joint features. The joint features are a concatenation of (1) the sum of image features along the spatial dimensions, weighted by $H^P_T$; and (2) the max-pooled image features along with the spatial dimension. We employ an MLP to process joint features, and further predict a discrete probability distribution $P^P_R$ for Euler angles (discretized into bins with a resolution of $5^\circ$) as well as other binary action variables $V^P_O$. For the primary arm action $a^P_{bc}$, we perform argmax on $H^P_T$ and $P^P_R$ to output it.

\subsection{Auxiliary Decoder}

As shown in Fig.~\ref{fig3}~(b), our auxiliary decoder is designed to predict the actions of the auxiliary arm by leveraging both global features and contextual information from the primary decoder. To distill essential information from the primary decoder outputs, we extract the 2D points $V^P_T$ across different views from $H^P_T$ and three Euler angles $V^P_R$ from $P^P_R$ with the highest score. These features are concatenated with $V^P_O$ to form a token encapsulating the most salient aspects of $a^P_{bc}$. After projection through an MLP, this token is concatenated with a learnable query embedding representing $a^A_{pc}$ to construct action tokens. For image/language tokens, we first apply an MLP to extract features relevant to the auxiliary arm's operational regions, preserving the primary arm's prediction integrity.
The decoder has one cross-attention layer and six self-attention layers. 
Due to the input of the primary arm action knowledge and the prediction in $C_{pc}$, the auxiliary decoder only needs to perform the cross-attention between action tokens and image/language tokens.
Since these tokens are much smaller than the image/language tokens, they can ignore unimportant areas in the image and retain the features of the key regions while reducing the computational cost. For the final output, these tokens pass through MLP to predict $a^A_{pc}$. Note that $a^A_{pc}$ contains 2D continuous points across different views $\hat{V}^A_T$ for translation, three continuous Euler angles $\hat{V}^A_R$ for rotation, and other action variables $V^A_O$.
\subsection{Role Assignment Module}
The structure of the proposed role assignment module is illustrated in Fig. \ref{fig3} (c). We initialize a learnable query embedding with random values to represent the role-specific latent variable $\xi$. The contextual information derived from both image and language plays a critical role in determining the functional division between the two arms. To achieve dynamic and context-aware role switching, we first compute cross-attention between the role query and the image/language tokens processed by MLP. This interaction enables the query to attend to the semantically salient regions and instructions. We further apply six cross-attention blocks between the role query and the action tokens, thereby aligning role assignment with intended motor actions. Finally, the fused representation is projected through an MLP to produce a binary classification variable, which explicitly dictates the arm assignment under the current perceptual and semantic context.

\begin{table*}[t]
\caption{The experimental result in simulation. We train all policies based on 10 or 100 training demonstrations, and evaluate on the same 25 episodes of the test set. As Kstar Diffuser \cite{Lv_2025_CVPR} has not released the code, we report its performance metrics from the original publication.}
\label{table1}
\centering
\begin{tabular}{ccccccccccccc}
    \hline
    & & & Push & Lift & Lift & Put in & Pick & Pick & Sweep & Handover & Put in & Take out \\
    Method & Architectural & Avg. & Box & Ball & Tray & Drawer & Plate & Laptop & Duspan & (easy) & Bridge & Tray \\
    \hline
    \hline
    & \multicolumn{11}{c}{\textbf{\textit{20 demos}}} \\
    \hline
    RVT-LF \cite{grotz2024peract2}  & Dual-model  & 2.4 & 12& 4 & 0 & 8  & 0 &  0 &  0 &  0 & 0 & 0 \\
    PerAct-LF \cite{grotz2024peract2} & Dual-model & 3.6 & 8 & 4 & 0 & 16  & 0  &  0 & 8  & 0 & 0 & 0\\
    AnyBimanual \cite{Lu2025anybimanual} & Dual-model & 11.6 & 28 & 4 & 4  & 12  & 12 & 8 & 16  & 24 &  0 & 8\\
    \hline
    ACT  \cite{zhaolearning}    & Single-model      & 4   & 0   & 24  & 4   & 12  & 0   & 0  & 0   & 0  & 0   & 0\\
    PerAct2 \cite{grotz2024peract2} & Single-model  & 4 &  0 & 8  &  0 & 12  & 4 & 0 & 8  & 8 & 0 & 0\\
    Kstar Diffuser \cite{Lv_2025_CVPR} & Single-model & -  & 80  & 87 & -   & -  & -   & 17 & 83  & 24 & -   & - \\
    \textbf{\name(ours)}   & Single-model  & \textbf{61.6}  & \textbf{84}   & \textbf{100}  & \textbf{80}   & \textbf{80}  & \textbf{24}   & \textbf{44} & \textbf{100}   & \textbf{36}  & \textbf{32}   & \textbf{36} \\
    \hline
    \hline
    & \multicolumn{11}{c}{\textbf{\textit{100 demos}}} \\
    \hline    
    RVT-LF \cite{grotz2024peract2}    & Dual-model  & 10 & 52 & 16 & 8  & 12 & 4  & 4  & 0  & 0  & 0  & 4 \\
    PerAct-LF  \cite{grotz2024peract2} & Dual-model & 20  & 56 & 40 & 16 & 28 & 4  & 12 & 28 & 8  & 0  & 8\\
    AnyBimanual \cite{Lu2025anybimanual} & Dual-model & 20    & 24 & 32 & 12 & 20  & 32 & 8  & 32 & 28 & 8  & 4\\
    \hline
    ACT     \cite{zhaolearning}   & Single-model  & 6 & 0  & 36 & 8 &  12 & 0  & 0  & 0  & 0  & 0  & 4\\
    PerAct2   \cite{grotz2024peract2}  & Single-model & 14  & 8  & 52 & 4  & 12 & 4  & 12 & 0  & 40 & 4  & 8\\
    Kstar Diffuser \cite{Lv_2025_CVPR} & Single-model & - & 83 & 98 & - & - & - & 44 & 89  & 27 & -   & -\\
    \textbf{\name(ours)} & Single-model  & \textbf{68.8} & \textbf{88} & \textbf{100} & \textbf{88} & \textbf{60} & \textbf{40} & \textbf{48} & \textbf{96} & \textbf{52} & \textbf{48} & \textbf{68}\\
    \hline
\end{tabular}
\end{table*}

\subsection{Coordinate Transformation}
$a^A_{pc}$ obtained from the auxiliary decoder is in the coordinate system $C_{pc}$. However, execution requires its representation in the base coordinate system $C_{bc}$. We therefore transform $a^A_{pc}$ from $C_{pc}$ to $C_{bc}$ through the following operations:
\begin{equation}\label{eq:rotation_transform}
V^A_R = \Gamma^{-1}\left( \Gamma(V^P_R) \cdot \Gamma(\hat{V}^A_R) \right)
\end{equation}
\begin{equation}\label{eq:translation_transform}
V^A_T = V^P_T + \hat{V}^A_T
\end{equation}
where $\Gamma$ \cite{mueller2019modern} denotes the Euler-angle-to-rotation-matrix transformation, $V^A_R$ represents the continuous Euler angle in $C_{bc}$, and $V^A_T$ corresponds to multi-view 2D points in $C_{bc}$.  This transformation yields the base-coordinate action $a^A_{bc}$. Noted that $V^A_T$ and $V^P_T$ are back-projected to the 3D point for translation in evaluation. Finally, $a^P_{bc}$ and $a^A_{bc}$ are mapped to the left/right arm according to $\xi$.

\subsection{Loss Function}
We train \name using a group of losses. For the primary arm action prediction, we compute cross-entropy losses for $H^P_T$, $P^P_R$, and $V^P_O$:
\begin{equation}
\mathcal{L}^P = CE(H^P_T,Y^P_T)+CE(P^P_R,Y^P_R)+CE(V^P_O,Y^P_O)
\end{equation}
where ground truth $Y^P_T$ denotes the 2D points across different views projected from the ground-truth 3D point of the primary arm, while $Y^P_R$ and $Y^P_O$ represent rotation, gripper state, and whether to allow collision avoidance.

For the auxiliary arm, we extract corresponding ground truths $Y^A_T$, $Y^A_R$, and $Y^A_O$ from datasets. Since $V^A_R$ is a periodic continuous variable on $[0^\circ, 360^\circ]$, the general mean squared error (MSE) fails, as shown in Fig. \ref{fig3} (d). To address this limitation, we introduce a circular MSE loss:
\begin{equation}
\mathcal{L}^A_{rot}=||\min(|V^A_R-Y^A_R|, 360-|V^A_R-Y^A_R|)||^2
\end{equation}
The auxiliary arm's total action loss combines:
\begin{equation}
\mathcal{L}^A=MSE(V^A_T,Y^A_T)+CE(V^A_O,Y^A_O)+\kappa\mathcal{L}^A_{rot}
\end{equation}
where the scaling factor $\kappa = 1/360$ (default) normalizes the magnitudes of $\mathcal{L}^A_{rot}$.
The role assignment loss is:
\begin{equation}
\mathcal{L}^\xi = CE(\xi, Y_\xi)
\end{equation}
where $Y_\xi$ represents the ground truth. Finally, the training loss of \name is as follows:
\begin{equation}
\mathcal{L}^{total} = \mathcal{L}^P + \lambda_A \mathcal{L}^A + \lambda_\xi \mathcal{L}^\xi
\end{equation}
with $\lambda_A$, $\lambda_\xi$ as task-balancing hyperparameters.

\section{Experiments}


\subsection{Experiment Settings}


\textbf{Simulation.} Bimanual manipulation tasks present significantly greater challenges than their single-arm counterparts due to stringent requirements for coordination, synchronization, and symmetry awareness between dual robotic arms. To evaluate the capabilities of \name in these complex regimes, we employ the RLBench2 \cite{grotz2024peract2} benchmark encompassing 10 distinct language-conditioned tasks. This suite explicitly includes synchronous, asynchronous, symmetric, and asymmetric manipulation scenarios. For training of \nameo, we implemented a $\xi$ annotation schema in the dataset to explicitly designate left/right arm roles (primary vs. auxiliary) for model guidance.
For environmental observation, a multi-camera system comprising six RGB-D cameras (resolution: 128 × 128) provides comprehensive coverage of the entire workspace. During policy training, each task is supported by 20 or 100 expert demonstrations. Evaluation rigor is ensured by executing 25 episodes per task within the RLBench2 \cite{grotz2024peract2} testing set, mitigating stochastic variance. 

\textbf{Real-World.} 
For real-world validation, we employ a dual-arm system comprising two Yahboom DoFbot manipulators and design two representative tasks: \texttt{handover} and \texttt{grasp banana}. Training data are collected using Moveit in ROS2 to control the robotic arms for demonstration recording. Observations are provided by a calibrated front-facing RGB-D camera capturing images at a resolution of 1280 × 720. We gather 15 demonstration trajectories per task. For evaluation, each task is executed over 10 trials using an NVIDIA RTX 4090 GPU.

\textbf{Baseline.} 
We evaluate \name against state-of-the-art bimanual manipulation methods, categorized as follows:

(1) Dual-model: RVT-Leader Following (RVT-LF) \cite{grotz2024peract2} employs an RVT \cite{goyal2023rvt} backbone with a leader-follower mechanism; Perceiver-Actor Leader Following (PerAct-LF) \cite{grotz2024peract2} applies a similar leader-follower paradigm using PerAct \cite{shridhar2023perceiver}; AnyBimanual \cite{Lu2025anybimanual} transfers pre-trained PerAct \cite{shridhar2023perceiver} via a skill manager and visual aligner module. 

(2) Single-model: Action Chunking with Transformers (ACT) \cite{zhaolearning} uses a Conditional VAE \cite{kingma2013auto} for joint angle sequence prediction; 
PerAct2 \cite{grotz2024peract2} enhances PerAct by unifying dual-arm action prediction within a shared feature space;
Kstar Diffuser \cite{Lv_2025_CVPR} is a generative model predicting kinematics-aware actions using a physics-grounded spatial-temporal graph to condition denoising.



\textbf{Implementation Details.} Following PerAct2 \cite{grotz2024peract2}, we implement $\mathrm{SE}(3)$ observation augmentation for expert training demonstrations to enhance model robustness. All comparative methods undergo standardized training: 100k iterations on NVIDIA A100 GPUs with a global batch size of 64. Model optimization employs the LAMB optimizer \cite{you2019large} with an initial learning rate of $5 \times 10^{-4}$, utilizing a cosine decay schedule with a linear warmup phase spanning 3k iterations.

\begin{table}[t]
\caption{Ablation Studies. We evaluated the influence of decoder, $C_{pc}$, and $\xi$ on performance across three representative tasks.}
\label{table2}
\centering
\resizebox{.98\columnwidth}{!}{
\begin{tabular}{ccc|ccc|c}
    \hline
     &  &  & Push & Put in  & Take out  &  \\
    Decoder & $C_{pc}$ & $\xi$ & Box & Drawer & Tray & Avg. \\
    \hline
    \hline
    \textbf{Primary-Auxiliary} & \textbf{Y} & \textbf{Y} & \textbf{88} & \textbf{60} & \textbf{68} & \textbf{72}\\
    Primary-Auxiliary & Y & N & 88 & 60 & 28 & 58.7\\
    Primary-Auxiliary & N & Y & 44 & 32 & 68 & 48\\
    Primary-Auxiliary & N & N & 44 & 32 & 28 & 34.7 \\
     Primary-Primary & N & N & 20 & 24 & 12 & 18.7 \\
    \hline
    
\end{tabular}}
\end{table}

\begin{table}[t]
\caption{The results in the real world. }
\label{table3}
\centering
\begin{tabular}{c|ccc}
    \hline
    Method & Avg. & Grasp Banana & Handover \\
    \hline
    \hline
    ACT \cite{zhaolearning} & 5 & 10 & 0 \\
    PerAct2 \cite{grotz2024peract2} & 35 & 30 & 40 \\
    \name & \textbf{85} & \textbf{90} & \textbf{80} \\
    \hline
\end{tabular}
\end{table}

\begin{figure*}[t]
\centering
\includegraphics[width=0.93\textwidth]{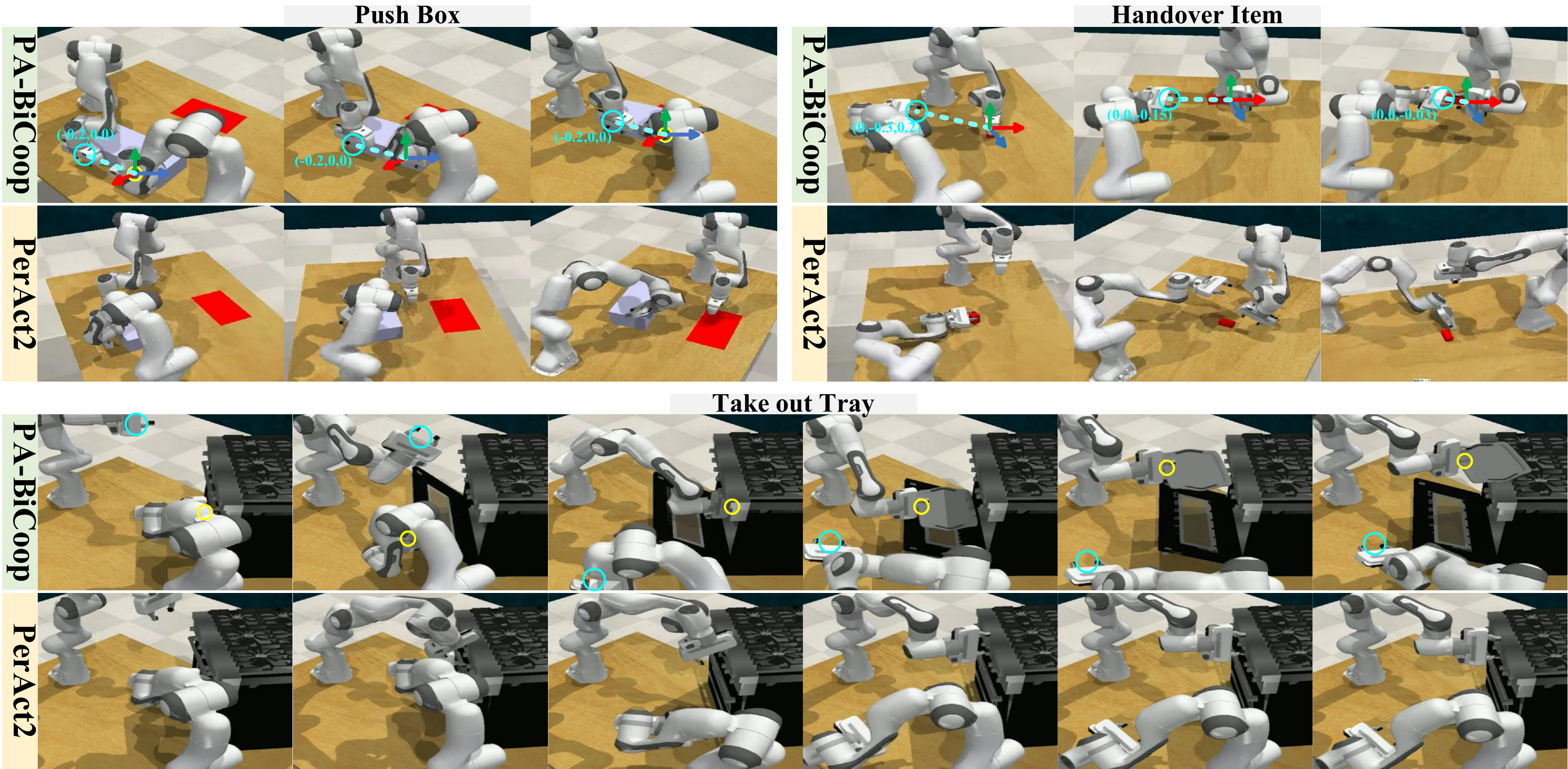} 
\caption{The visualization on RLBench2. The yellow circles represent the primary arm actions, the blue circles represent auxiliary arm actions, and the vector beside the blue circle is the translation vector of the auxiliary arm in $C_{pc}$.}
\label{fig4}
\vspace{-0.20cm}
\end{figure*}

\subsection{Simulation Results}
TABLE \ref{table1} presents comparative results between \name and state-of-the-art methods under 20 and 100 demonstration training regimes. \name achieves an overall performance improvement of at least $48\%$ against all baselines. We observe that:

(1) For symmetrical and synchronous tasks (e.g., \texttt{push box}, \texttt{lift ball}, \texttt{lift tray}), temporal misalignment induces insufficient force application or object instability, resulting in task failure. Our relative coordinate system prediction mechanism ensures operational stability, achieving success rates exceeding $84\%$. In the \texttt{lift ball}, we even achieved a $100\%$ success rate.

(2) For asynchronous and asymmetric tasks such as \texttt{put in drawer} and \texttt{handover (easy)}, where bimanual coordination complexity increases, \name maintains at least $4\%$ superiority over baselines despite performance attenuation. In the \texttt{put in drawer} task, the auxiliary arm dynamically determines which drawer to open based on linguistic instructions. Despite this requirement, \name achieves a success rate of at least $60\%$, demonstrating our auxiliary decoder's strength in spatial relationship perception.

(3) Furthermore, \name demonstrates substantial advancement in long-horizon tasks (\texttt{put in fridge}, \texttt{take out tray}), elevating success rates from under $10\%$ to over $45\%$ based on 100 training demonstrations. This performance leap is directly attributable to our observation-driven role switching mechanism.

Across all evaluated tasks, Kstar Diffuser \cite{Lv_2025_CVPR} and PerAct2 \cite{grotz2024peract2} demonstrate significantly lower success rates than \name despite employing a comparable single shared model for bimanual motion prediction. This performance gap originates from their treatment of the left and right arms as functionally equivalent, without distinguishing their respective roles, thereby failing to capture the critical aspects of bimanual manipulation. These results validate the effectiveness of \name in achieving synergistic bimanual coordination.
\begin{figure*}[t]
\centering
\includegraphics[width=1.0\textwidth]{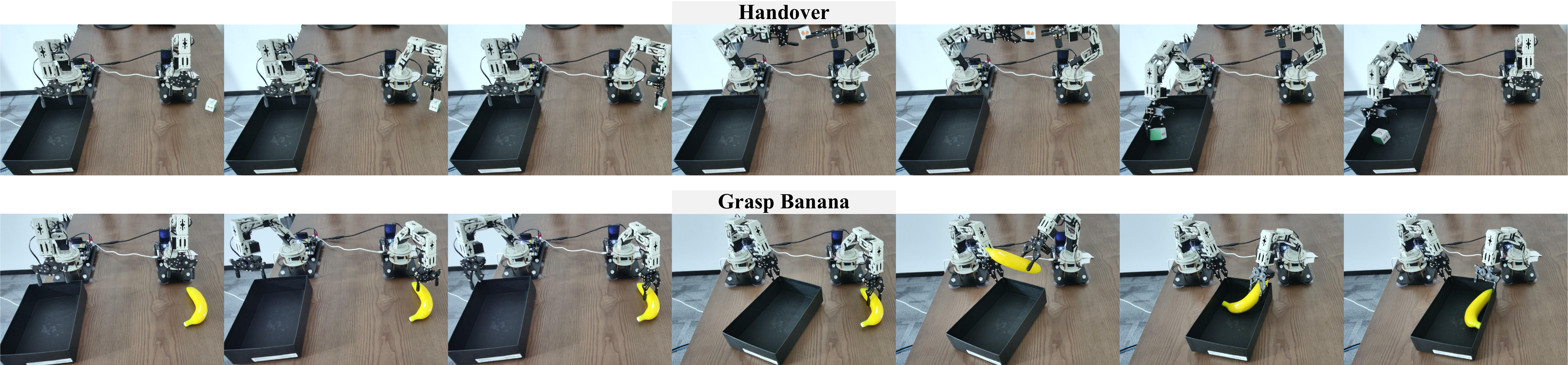} 
\caption{The visualization of our \name in real-world tasks using two Yahboom DoFbot manipulators. }
\label{fig5}
\end{figure*}

\subsection{Ablation Studies}

In this section, we evaluate the core proposed components of our \name via ablation studies in TABLE \ref{table2} on representative tasks: \texttt{push box}, \texttt{put in drawer}, and \texttt{take out tray}.

\textbf{Effects of primary-auxiliary.}
We designed a primary-primary architecture, in which each arm’s actions were processed independently by a dedicated primary decoder, to validate the efficacy of our primary-auxiliary cooperative framework.
Due to the absence of role specialization and knowledge sharing between the arms, this model exhibits a significant decline in performance, with the average success rate across various tasks dropping by more than $16\%$.

\textbf{Effects of $C_{pc}$.} 
Predicting the auxiliary arm pose in the coordinate system $C_{bc}$ results in a $24\%$ reduction in average success rates. This performance degradation occurs because the auxiliary arm pose exhibits substantial variations in this coordinate system, which substantially increases the learning complexity of the auxiliary arm pose estimation. Conversely, our \name predicts the auxiliary arm pose in $C_{pc}$, where the pose is defined relative to the precise primary arm pose and remains invariant to manipulated object displacements. This stability proves particularly beneficial in synchronous tasks, where constant spatial relationships between the primary and auxiliary arms significantly reduce bimanual coordination complexity. Our ablation study conclusively demonstrates the efficacy of the design $C_{pc}$ for coordinated bimanual manipulation.

\textbf{Effects of $\xi$.}
Our experiments reveal that in the long-horizon \texttt{take out tray} task, success rate declines to $28\%$ without observation-driven role switching between arms. This performance degradation occurs because restricting the primary role to a single arm during extended tasks reduces manipulation precision for the other arm operation. Consequently, our ablation studies validate the effectiveness of $\xi$ in resolving this limitation.

\subsection{Qualitative Analysis}
As shown in Fig. \ref{fig4}, we evaluate \name against PerAct2 \cite{grotz2024peract2} across RLBench2 \cite{grotz2024peract2} benchmarks, conducting detailed qualitative assessments on synchronous, asynchronous, and long-horizon tasks.

In the \texttt{push box} task, both arms can serve as the primary arm. Crucially, the relative pose between arms remains constant throughout pushing the box. The action prediction of the auxiliary arm in $C_{pc}$ enables it to have the same high accuracy as the action prediction of the primary arm, demonstrating exceptional robustness and environmental adaptability. Conversely, PerAct2's \cite{grotz2024peract2} requires both arms to perform independent object recognition, preventing effective synchronization and consequently achieving suboptimal success rates.


In the \texttt{handover item} task, the arm nearer to the target item is designated as the primary arm, while the other arm is the auxiliary arm to hand over the item. Equipped with precise detection and positioning capabilities, our primary decoder enables the primary arm to accurately grasp the object. During handover, the auxiliary arm operates without the need for repeated precise object detection. Instead, it simply approaches the origin of $C_{pc}$ in $C_{pc}$, thereby significantly reducing the coordination complexity between the two arms. In contrast, PerAct2 requires both arms to perform precise object detection and localization, which substantially increases the computational burden on the model. Even minor inaccuracies in this process can lead to failures in the \texttt{handover item} task.

In the \texttt{take out tray} task, \name dynamically adapts role between arms: designating the right arm as the primary arm during oven-door operation, then switching primary role to the left arm for tray extraction. This adaptive coordination enables seamless bimanual cooperation. PerAct2 \cite{grotz2024peract2} fails to model such coordination mechanisms, resulting in significantly degraded long-horizon task performance.


\subsection{Real-World Results}

To further validate the effectiveness of our \name, we evaluate it on two real-world tasks, both requiring spatiotemporal coordination between the arms. Performance results are summarized in TABLE \ref{table3}. Despite the limited number of demonstrations in the training datasets, our \name outperforms existing approaches by over $50\%$. Although PerAct2 \cite{grotz2024peract2} and ACT \cite{zhaolearning} also employ a single shared model, they do not differentiate the roles of the two arms and exhibit limited capability in tasks demanding precise bimanual coordination.
Qualitative results of \name on both tasks are visualized in Fig. \ref{fig5}. The \texttt{handover} and \texttt{grasp banana} tasks involve placing objects into a fixed or movable box, and both arms must operate in coordination due to the limitation of spatial distance. Our \name facilitates knowledge exchange between arms, dynamically assigns complementary roles, and achieves a high degree of coordination, leading to successful task completion.

\section{Conclusion}
This paper addresses key challenges in bimanual manipulation, where previous models lack inter-arm interaction and ignore the dynamic division of labor. Drawing on human bimanual mechanisms, we propose PA-BiCoop, a single-model framework with dynamic primary-auxiliary arm differentiation: it includes two specialized decoders that share a global encoder, the primary decoder generates the primary arm’s base-coordinate pose and core-task affordance heatmaps, while the auxiliary decoder outputs the auxiliary arm’s relative pose. It also has a dynamic role assignment module that automatically maps primary or auxiliary roles to left and right arms without manual pre-definition. Extensive experiments demonstrate PA-BiCoop’s efficacy in complex bimanual manipulation, confirming it enhances coordination efficiency and adaptability and lays a foundation for robust robotic bimanual systems.

\textbf{Limitations and prospects for the future.} Currently, PA-BiCoop faces difficulties in extremely long-horizon tasks, e.g., multi-step assembly involving dozens of sequential operations or tasks with extended pauses. In future work, we will aim to optimize the role assignment module by incorporating task stage prediction capabilities or introducing a lightweight task memory mechanism.

\bibliography{bib/IEEEabrv,bib/ref}
\end{document}